\documentclass[journal]{IEEEtran}
\IEEEoverridecommandlockouts
\usepackage{cite}
\usepackage{amsmath,amssymb,amsfonts}
\usepackage{algorithmic}

\usepackage{graphicx}
\usepackage{subfigure}

\usepackage{multirow}
\usepackage{float}
\usepackage{bbding}

\usepackage{mathtools}
\usepackage{mathrsfs,array,color,amsthm}
\usepackage{bm} % bold

\usepackage{epstopdf}
\usepackage{booktabs}
\usepackage{textcomp}
\usepackage{xcolor}
\usepackage[linewidth=1pt]{mdframed}
\def\BibTeX{{\rm B\kern-.05em{\sc i\kern-.025em b}\kern-.08em
    T\kern-.1667em\lower.7ex\hbox{E}\kern-.125emX}}

\begin{document}

\title{Multiscale Convolutional Transformer with Center Mask Pretraining for Hyperspectral Image Classification}

\author{Sen Jia, Yifan Wang}

\maketitle

\begin{abstract}
Hyperspectral images (HSI) not only have a broad macroscopic field of view but also contain rich spectral information, and the types of surface objects can be identified through spectral information, which is one of the main applications in hyperspectral image related research.In recent years, more and more deep learning methods have been proposed, among which convolutional neural networks (CNN) are the most influential. However, CNN-based methods are difficult to capture long-range dependencies, and also require a large amount of labeled data for model training.Besides, most of the self-supervised training methods in the field of HSI classification are based on the reconstruction of input samples, and it is difficult to achieve effective use of unlabeled samples. To address the shortcomings of CNN networks, we propose a noval multi-scale convolutional embedding module for HSI to realize effective extraction of spatial-spectral information, which can be better combined with Transformer network.In order to make more efficient use of unlabeled data, we propose a new self-supervised pretask. Similar to Mask autoencoder, but our pre-training method only masks the corresponding token of the central pixel in the encoder, and inputs the remaining token into the decoder to reconstruct the spectral information of the central pixel.Such a pretask can better model the relationship between the central feature and the domain feature, and obtain more stable training results.
\end{abstract}

% \fi
\IEEEpeerreviewmaketitle

\section{Introduction}
Hyperspectral images, captured by different types of satellites, are generally composed of dozens to hundreds of bands and have the characteristics of low spatial resolution and high spectral resolution. A hyperspectral image usually provides a good overview of macroscopic surface features. The spectral information contained in the image provides the possibility to distinguish land covers, which also makes hyperspectral images widely used in various fields.This has spawned various practical applications, such as agricultural industry \cite{agricultural}, urban planning \cite{urban}, change detection \cite{change}, Lesion detection \cite{medical} and mineral detection \cite{mineral}, etc. \par

Among them, pixel-level classification of HSI is the most concerned one in the community, and its main task is to assign a class label to each pixel, which is somewhat like semantic segmentation in the computer vision (CV) field. Due to the high spectral resolution of hyperspectral images, there must be redundant bands. Principal component analysis \cite{PCA} and independent component analysis \cite{ica} are widely used for redundancy elimination and have achieved good results.In the early stage of research, people mostly combine manual feature extraction methods with traditional classifiers, such as Logistic regression \cite{logistics}, decision tree \cite{tree}, random forest \cite{random}, and SVM \cite{svm} to classify the ground objects by spectral information.However, the imaging distance of HSI is distant and there are many inference factors in this process. Therefore, the spectral curve of different surface objects is not always easy to distinguish. This is a problem for these classification methods that take spectral information as input, and they cannot capture the patterns between features well.In recent years, deep learning methods have gradually become popular in HSI classification tasks, in which CNN-based methods are dominant.In \cite{1dcnn}, Hu et al. made a preliminary attempt of CNN models in the hyperspectral image classification task that several 1d convolutional layers are stacked to extract local spectral information, and several data augmentation method in CV field has been introduced. In \cite{hybrid}, the authors combine 3D convolution and 2D convolution to achieve hierarchical feature learning. \cite{3dcnn} Combining 3D convolution with 1D convolution reduces the computational cost of 3D convolution. In addition, other neural networks have also achieved good performance. Zhou et al. \cite{sslstm} designed two-branch Long Short-Term Memory network (LSTM) to extract spectral information and spatial information respectively. He et al. \cite{mlp} proposed a pure mlp structure, proving that the mlp network still has potential. In \cite{graph}, Hong et al. designed a mini batch graph neural network for HSI. It is worth mentioning that the recently very popular Transformer model has also been introduced into the field of hyperspectral classification to improve the long-range modeling capability of CNN-based model. Hu et al. \cite{1dtrans} uses 1D convolution as an embedding layer combined with Transformer Encoder and introduces a metric learning loss. Zhong et al. \cite{sstn} proposed an efficient and effective spectral–spatial transformer network and a model structure search framework. \par

However, deep learning models often require a large number of labeled samples as training samples, but the labeled data of HSI is limited, and large-scale sample labeling is obviously time-consuming and labor-intensive.Recently, self-supervised learning has developed vigorously. Autoregressive language modeling in GPT \cite{GPT} and masked autoencoding in BERT \cite{BERT} are one of the most representative methods in natural language processing (NLP). Self-supervised methods in computer vision (CV) are mainly dominated by contrastive learning methods, such as Momentum Contrast (MoCo) series \cite{mocov1, mocov2, mocov3} and SimCLR series \cite{simclrv1, simclrv2}. Inspire by BERT, MAE \cite{MAE} leads a new hotspot of self-supervised methods in CV, and its mask-reconstruction pretask is also very instructive for related research in HSI classification field. 

In this article, a multiscale convolutional transformer is proposed for HSI classification, and a new self-superviesd pretask is designed. The main contribution are listed as follows:

\begin{itemize}  %% 无数字标记的清单项
	\item According to the characteristics of hyperspectral data, a new unsupervised pre-training pretask is proposed. By the pretask of center pixel reconstructing, the model's ability to capture the relationship between the center pixel and surrounding pixels and the semantic understanding of the sample are improved. 
	\item In conjunction with our proposed pre-training strategy, we propose a multi-scale convolutional Transformer as the backbone network. We replace the commonly used linear embedding module with a well-designed multi-scale convolutional embedding module, combined with a spectral segmentation strategy, to achieve efficient spatial- spectral feature embedding. 
	\item A series of comparative experiments and ablation experiments demonstrate the effectiveness of our proposed pre-training method and backbone network. In particular, our proposed pre-training method can well eliminate the instability of results caused by random sampling in limited training samples scenario.
\end{itemize}\par
The rest of this paper is organized as follows. Our proposed method will be introduced in detail in section \uppercase\expandafter{\romannumeral2}. The descriptions of result analysis and comparative experiment will be provided in section \uppercase\expandafter{\romannumeral3}, and section \uppercase\expandafter{\romannumeral4} presents the conclusions.

\section{Methodology}

% BackBone network
\begin{figure*}[!t]
    \centering
    \includegraphics[scale = 0.25]{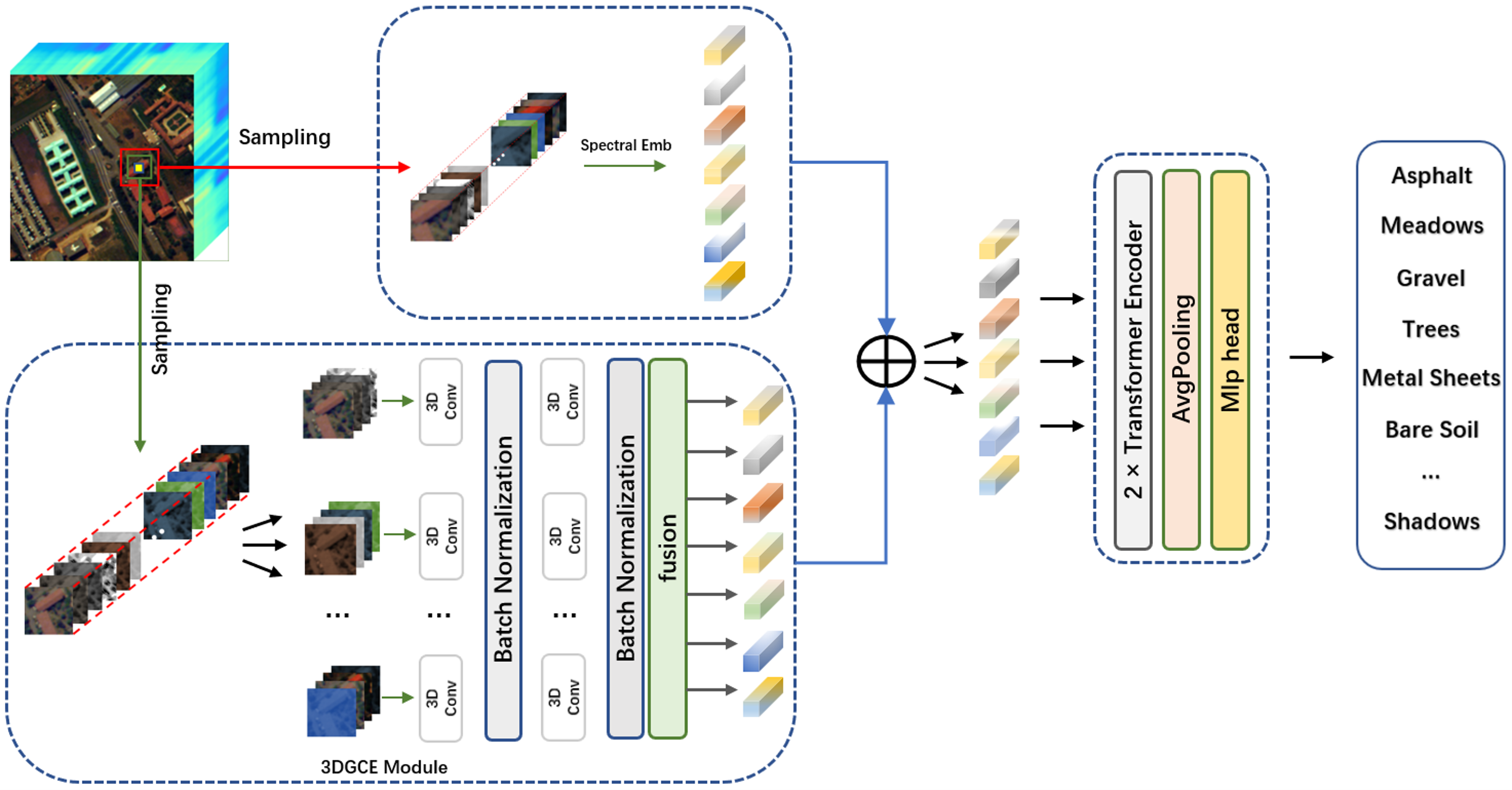}
\caption{Architecture of the proposed multiscale convolutional transformer for hyperspectral image classification}
\label{fig:BackBone}
\end{figure*}

% pretask 
\subsection{Selfsuperviesd Learning}
\begin{figure*}[!t]
    \centering
    \includegraphics[scale = 0.2]{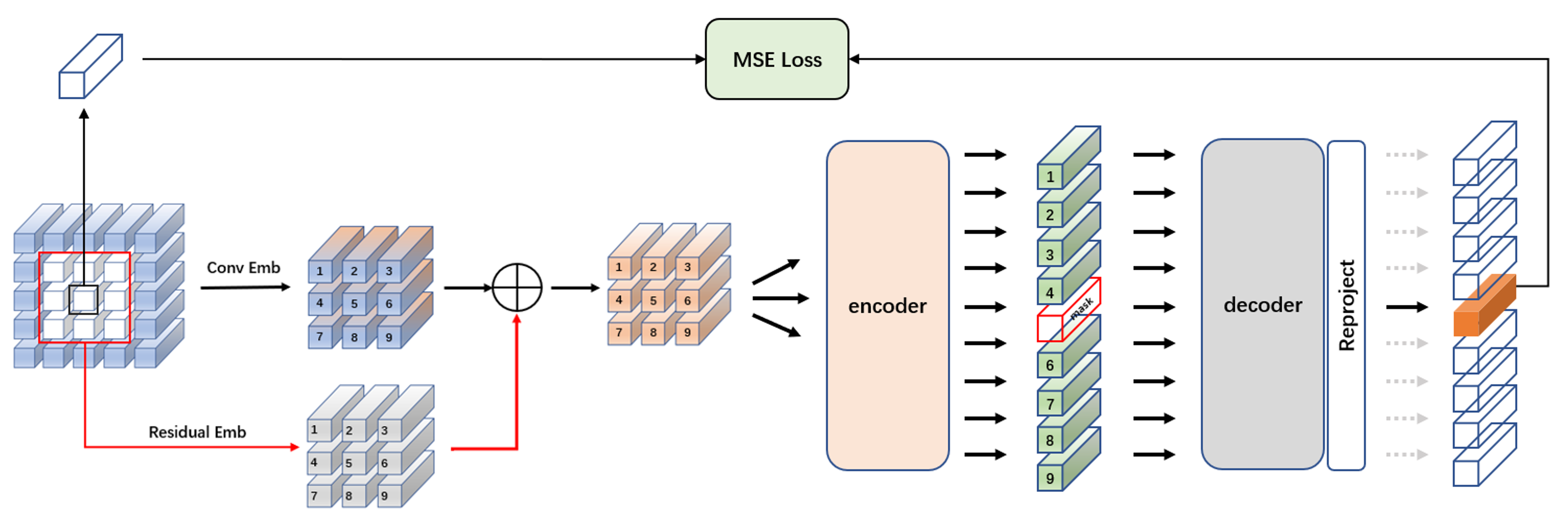}
\caption{Architecture of the proposed center mask pretraining pretask}
\label{fig:pretask}
\end{figure*}

In this section, I will introduce our proposed Multiscale convolutional Transformer (MCT) network and Center-mask pre-training pretask (CMPP).  Detailed flowchart of the proposed method is in \ref{fig:BackBone} and \ref{fig:pretask}. The MCT network has three parts, the multiscale convolutional embedding (MCE) module, Transformer Encoder (TE) , and MLP head. And CPMAE, through the mask-reconstruction process of the center pixel of the input patch, enables the backbone network to effectively model the center pixel and neighborhood pixel relationship in the self-supervised learning process.

\subsection{Multiscale Convolutional Transformer}

\subsubsection{Multiscale Convolutional Embedding}
Considering that the Vit Transformer network was originally used for RGB images, and the hyperspectral data has low spatial resolution and high spectral resolution, the original linear embedding and image segmentation strategies cannot meet the characteristics of hyperspectral data. So we propose the Multiscale Convolutional Embedding Module, which consists of a 3D spectral partition convolutional embedding (3DSPCE) branch and an independent information embedding (IIE) branch. In 3DSPCE branch, we combine the spectral partition strategy with 3D convolution. Because the spectral curves of objects often have local differences, It is difficult for a 3D convolution kernel to effectively extract the information of the entire band. Therefore, we employ a spectral partition strategy to divide the spectrum into several subbands of equal length. 3D convolution extraction is performed for each subband, and the same convolution operation is performed twice.In addition, to unify the feature deviations generated by different spectral segments after 3D convolution, we added a 3D BatchNormalization operation. Finally, we concatenate the features, and then complete feature fusion and compression. In particular, spectral segmentation operations have a simple implementation, and the convolution function often includes a grouped convolution option. \par
However, only using 3D convolutional embedding can only indiscriminately extract local features of surface objects, but cannot highlight the characteristics of central features. This makes it difficult to better model the relationship between land covers. To improve this problem, we propose the independent information embedding branch, which passes each pixel of the original image through a linear embedding layer and fuses the result of the convolutional embedding by skip connections. In this way, our input sequence not only contains the information of the ground objects themselves, but also the information of the neighboring ground objects, and such multiscale embedded features are more discriminative and robust. It is worth mentioning that the convolution operation without padding will change the shape of the feature, and the shape of the IIE branch is consistent with the output of the convolution embedding branch.

\subsubsection{Transformer Encoder}

\begin{figure}[!t]
    \centering
    \includegraphics[scale = 0.4]{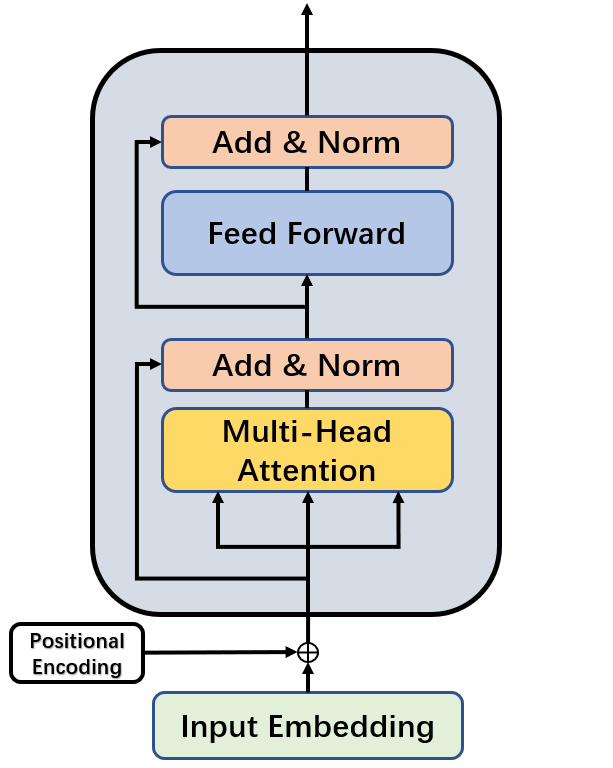}
\caption{Detail structure of a standard Transformer Encoder}
\label{transformer}
\end{figure}

The global perception ability of CNN often needs to make the model go deeper, but HSI data is limited so it is difficult for us to stack modules as simply as a computer vision model. While, The multi-head attention module can make up for the shortcomings of CNN here, and effectively model the relationship between ground objects. A standard Transformer Encoder is shown in \cite{fig:transformer}, which is mainly composed of position encoding, multi-head attention, and feedforward layers. Considering that the convolution operation itself contains position information, the position encoding is not used here, and the embedded multiscale features are directly input into the module. The definition of Scaled Dot-Product Attention and multi-head attention \cite{transformer} are as follows:
\begin{equation}
\label{eq2}
\begin{aligned}
      {\rm MultiHead}(Q, K, V)={\rm Concat}(head_{1},...,head_{h})W^O.\\
      {\rm where \;} head_{i} = {\rm Attention}(QW^Q_i, KW^K_i, VW^V_i).\\
\end{aligned}
\end{equation} 
 \begin{equation}
\label{fig:Transformer}
\begin{aligned}
      {\rm Attention}(Q, K, V)={\rm softmax}(\frac{QK^{\top}}{\sqrt{d}})V.
\end{aligned}
\end{equation} 
where $W^Q_i\in\mathbb{R}^{d_{models}\times d_k}$,$W^K_i\in\mathbb{R}^{d_{models}\times d_k}$,
$W^Q_i\in\mathbb{R}^{d_{models}\times d_v}$ and $W^O\in\mathbb{R}^{d_{models}\times hd_v}$ are parameter matrices.$\bm{Q}$ , $\bm{K}$ and $\bm{V}$ are the learnable weight matrices. $d_{model}$ and $d_v$ are the dimension of input embedding and $\bm{K}$. $d_k$ is the dimension of $\bm{Q}$ and $\bm{K}$. Finally, we add an average pooling layer to achieve feature compression and de-redundancy, and output the classification results through the three-layer MLP head.

\subsection{Center-mask Pre-training Pretask}
Today, most hyperspectral image classification methods are patch-based, that is, the input of the model is not only the spectral curve of center pixel, but also its neighbor region, which are generally square areas, so that the input samples can be more distinctive. Inspired by the form of the HSI sample, we propose the Center Mask (CM) Pre-training pretask, which is similar to MAE and adopts an asymmetric structure but is simpler and easier to implement. \par

The flowchart of CM Pre-training pretask is shown in \ref{fig:pretask}. The Encoder is MCT network (with the mlp header removed). The decoder consists of two layers of standard Transformer Encoders, which are only used in the pre-training process. Given an input sample $\bm{X}$, center pixel vector $\bm{V_c}$, The latent representation of the input sample is $\bm{E}$ (embedded by the MCT module). Unlike self-supervised pre-training in CV field, RGB images cannot directly find areas that need to be focused, but the neighborhood areas of HSI samples are to enrich the spatial features of the central pixel. In this way, our mask target can select the most important part of the input, that is, the center pixel. Therefore, we use a learnable vector $\bm{V_l}$ to replace the representation corresponding to the center pixel without disturbing the order of the embedding sequence. Then, the embedded sequence is input into the decoder, and the pixel-level reconstruction is performed by the MLP head to obtain the reconstruction result $\hat{\bm{V_l}}$ of the center pixel. The target of the CM pre-training task is to reconstruct the center pixel as efficiently as possible, so that the encoder can better learn the relationship between the center pixel and the neighbor pixels without labels. The reconstruction target can be formulated as: 
    \begin{equation}
        \mathcal{T}(V_c, \hat{V_l})=\min |V_c-\hat{V_l}|_{2}
    \end{equation}
where $\bm{T}$ is the similarity function. In the deep learning framework, Func $\bm{T}$ is equivalent to the mean squared error (MSE) loss function.

\section{Experiment}
To fully evaluate our proposed pre-training method and backbone network, we conduct comparative experiments and ablation experiments on two public datasets, Salinas and Yellow River Estuary (YRE). The false-color image and ground-truth are shown in Fig. 3 and Fig. 4 and The detailed information and The portion of the training set and test set are shown in the table. We use three metrics to evaluate the classification results, overall accuracy (OA), classwise average accuracy (AA), and kappa coefficient.

\subsection{Datasets Description}
\subsubsection{Salinas Dataset}
The salinas dataset, collected by the AVIRS sensor in the Salinas Valley, USA, has an image resolution of 512 × 217 and a spatial resolution of 3.7 meters. After noise band removal, 204 bands are remaining. There are 16 kinds of ground objects in the dataset, with a total of 56,975 samples that can be used for pixel-level classification.

% Salinas
\begin{table}[!t]
  \centering
  \caption{Land cover classes of the Salinas dataset, with the standard training and testing sets for each class.}
  \setlength{\tabcolsep}{3mm}{
  \renewcommand\arraystretch{1.5}
  \begin{tabular}{c|ccc}
  \toprule[1.5pt]
  Class&Class Name&Training&Testing\\
  \hline 
   1&Brocoli green weeds 1&5	&2004 \\
   2&Brocoli green weeds 2&5	&3721\\
   3&Fallow&5	&1971\\
   4&Fallow rough plow	&5	&1389\\
   5&Fallow smooth&5	&2673\\
   6&Stubble&5	&3954\\
   7&Celery&5	&3574\\
   8&Grapes untrained&5	&11266\\
   9&Soil vinyard develop	&5	&6198\\
   10&Corn senesced green weeds&5	&3273\\
   11&Lettuce romaine 4wk&5	&1063\\
   12&Lettuce romaine 5wk&5	&1922\\
   13&Lettuce romaine 6wk&5	&911\\
   14&Lettuce romaine 7wk&5	&1065\\
   15&Vinyard untrained&5	&7263\\
   16&Vinyard vertical trellis&5 &1802\\
  \hline  &Total &80 &54129\\
  \bottomrule[1.5pt]
  \end{tabular}}
  \label{Table:Sa-detail}
  \end{table}

\subsubsection{YRE Dataset}
YRE dataset is a large scene dataset captured by the Gaofen-5 satellite in the Yellow River Estuary region of Shandong Province, China. Its height and width are 1400, and the spatial resolution of each pixel is 30m, leaving 180 bands after removing noise bands. The surface objects are mainly wetland vegetation, there are 20 kinds of objects and the total number of labeled samples is 77,937.

% GF5
\begin{table}[!t]
  \centering
    \caption{Land cover classes of the YRE dataset, with the standard training and testing sets for each class.}
  \setlength{\tabcolsep}{3mm}{
  \renewcommand\arraystretch{1.5}
  \begin{tabular}{c|ccc}
  \toprule[1.5pt]
  Class&Class Name&Training&Testing\\
  \hline 
  1 & Building & 10 & 523 \\ 
  2 & River & 10 & 5366 \\ 
  3 & Salt Marsh & 10 & 4985 \\ 
  4 & Shallow Sea & 10 & 17540 \\ 
  5 & Deep Sea & 10 & 18667 \\ 
  6 & Intertidal Saltwater Marsh & 10 & 2333 \\ 
  7 & Tidal Flat & 10 & 1782 \\ 
  8 & Pond & 10 & 1777 \\ 
  9 & Sorghum & 10 & 636 \\ 
  10 & Corn & 10 & 1499 \\ 
  11 & Lotus Root & 10 & 2709 \\ 
  12 & Aquaculture & 10 & 8009 \\ 
  13 & Rice & 10 & 5498 \\ 
  14 & Tamarix Chinensis & 10 & 1210 \\ 
  15 & Freshwater Herbaceous Marsh & 10 & 1407 \\ 
  16 & Suaeda Salsa & 10 & 864 \\ 
  17 & Spartina Alterniflora & 10 & 570 \\ 
  18 & Reed & 10 & 1960 \\ 
  19 & Floodplain & 10 & 337 \\ 
  20 & Locus & 10 & 65 \\ 
  \hline  &Total &200 &77737\\
  \bottomrule[1.5pt]
  \end{tabular}}
  \label{Table:GF5-detail}
\end{table}

\subsection{Comparative Experiment}
To demonstrate the superiority of our proposed method, we select six state-of-the-art methods on two public datasets,Salinas and YRE, including five CNN methods and one classical Transformer network. They are CNNHSI \cite{cnnhsi}, SPRN \cite{sprn}, FC3D  \cite{fc3d}, HybridSN \cite{hybridsn}, TwoCNN \cite{twocnn} and Vision Transformers (ViT) \cite{vit}. Among them, several methods based on 2DCNN are distinguished in the size of the convolution kernel and the structure design. CNNHSI stacks several 2-D Convolution layer with 1 × 1 kernel size. SPRN adopts spectral partition (SP) operation and applies several parallel 3×3 convolutions. TwoCNN is a dual-branch CNN with a 2DCNN and a 1DCNN to extract spatial information and spectral information respectively. FC3D is a pure 3DCNN network, and HybridSN uses 3D convolution and 2D convolution successively for hierarchical feature extraction. Vit takes each pixel vector in the patch as the input. The model structure and parameter settings of the comparison methods comply with open source codes or corresponding papers. \par

% ---------------------------------   Sa fig  ---------------------------------
\begin{figure}[!t]
\centering

\subfigure[]
{
    \begin{minipage}[b]{.25\linewidth}
        \centering
        \includegraphics[scale = 0.4]{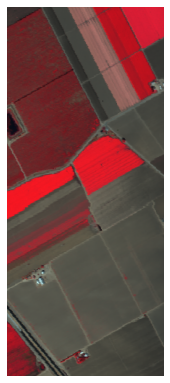}
    \end{minipage}
}
\subfigure[]
{
 	\begin{minipage}[b]{.25\linewidth}
        \centering
        \includegraphics[scale = 0.4]{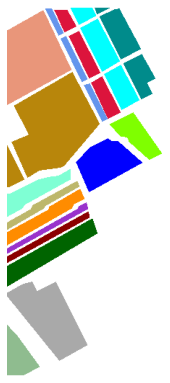}
    \end{minipage}
}
\subfigure
{
 	\begin{minipage}[b]{.25\linewidth}
        \centering
        \includegraphics[scale = 0.3]{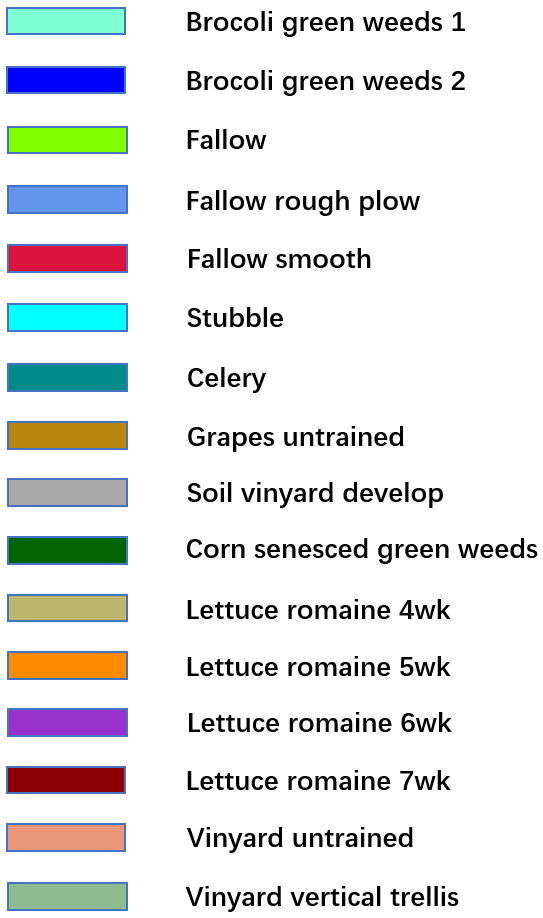}
    \end{minipage}
}\hspace{5mm}
\caption{(a) False-color image and (b) Ground-truth map of the Salinas dataset.}
\label{fig:Salinas}
\end{figure}

% ------------------------------- YRE fig --------------------------
\begin{figure}[htbp]

\centering

\subfigure[]
{
    \begin{minipage}[b]{.4\linewidth}
        \centering
        \includegraphics[scale = 0.3]{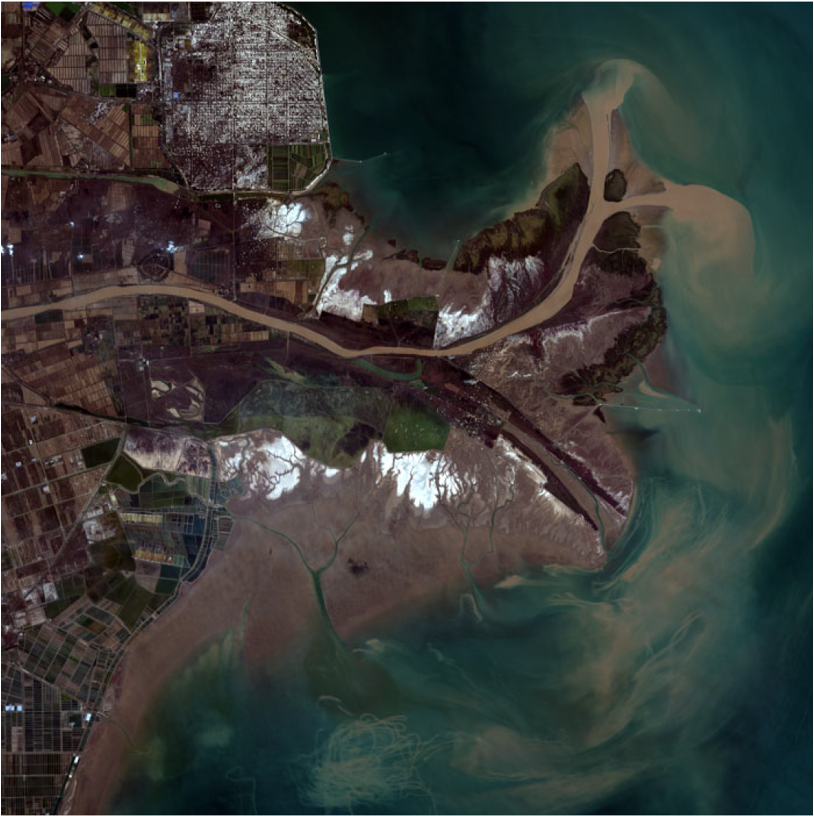}
    \end{minipage}
}
\subfigure[]
{
 	\begin{minipage}[b]{.4\linewidth}
        \centering
        \includegraphics[scale = 0.3]{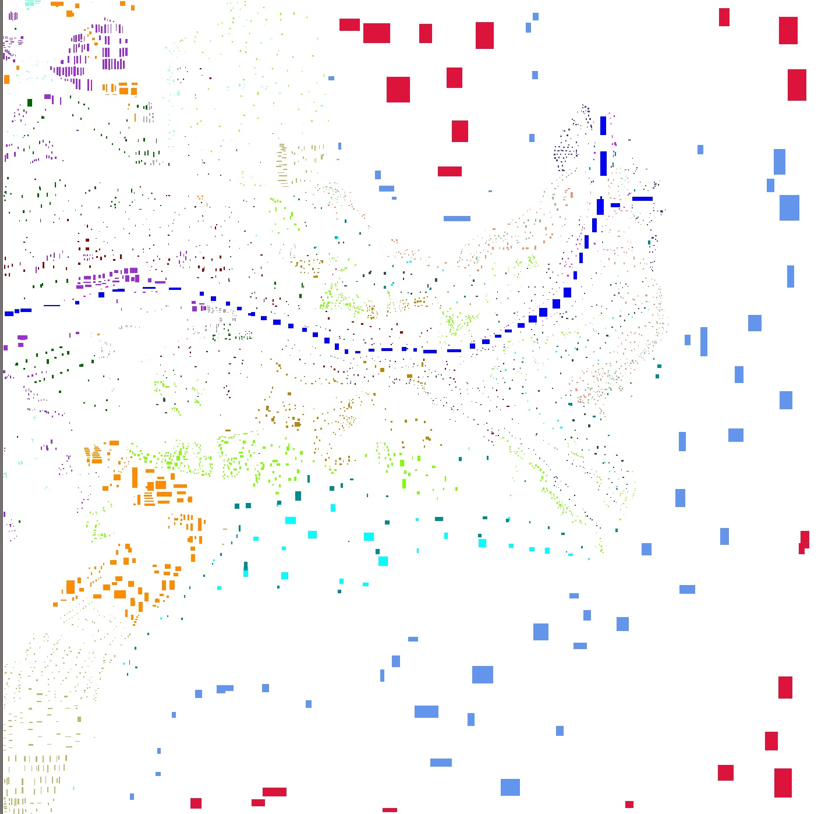}
    \end{minipage}
}
\caption{(a) False-color image and (b) Ground-truth map of the YRE dataset.}
\label{fig:Salinas}
\end{figure}

% SALINAS DATASET
\begin{table}[!t]
  \centering
  \caption{CLASSIFICATION ACCURACY (\%) AND KAPPA MEASURE FOR THE SALINAS DATASET}
  \setlength{\tabcolsep}{1.2mm}{
  \renewcommand\arraystretch{1.5}
  \begin{tabular}{c|ccccccc}
  \toprule[1.5pt]
  Class&Vit &HybridSN &FC3D &CNNHSI &TwoCNN &SPRN &MCT\\
  \hline  
  1 & \bf100.00  & 99.93  & 99.95  & 93.64  & 93.44  & 99.75  &\bf100.00  \\ 
  2 & 99.54  & 98.31  & 92.92  & 79.30  & 89.86  & 96.17  &\bf99.61  \\ 
  3 & 99.98  & 97.04  & 97.84  & 84.62  & 90.63  & 84.85  &\bf99.99  \\ 
  4 & 95.68  & 93.18  & 96.66  & 99.40  & 97.45  & 92.80  &\bf98.91  \\ 
  5 & 96.04  &\bf98.08  & 86.54  & 77.31  & 95.23  & 90.10  &\bf95.51  \\ 
  6 & 99.76  & 98.46  & 94.08  & 98.45  & 99.74  & 99.69  &\bf99.93  \\ 
  7 & 99.99  & 99.76  & 99.78  & 99.13  &\bf100.00  & 97.84  &\bf100.00  \\ 
  8 & 78.79  & 59.81  & 53.82  & 65.56  & 74.59  & 64.14  &\bf84.08  \\ 
  9 & 98.22  & 96.84  & 95.46  & 94.70  &\bf99.67  & 95.91  & 99.57  \\ 
  10 & 92.63  & 93.33  & 90.32  & 46.59  & 95.17  & 82.94  &\bf95.92  \\ 
  11 & 98.42  & 97.65  & 89.21  & 90.40  & 95.71  & 91.56  &\bf99.94  \\ 
  12 & 97.60  & 94.88  & 91.28  & 97.85  &\bf99.20  & 92.54  & 98.72  \\ 
  13 & 85.03  & 84.63  & 87.82  & 99.19  &\bf99.28  & 88.80  & 98.57  \\ 
  14 & 97.60  & 98.97  & 92.99  & 92.94  & 96.39  & 90.14  &\bf99.58  \\ 
  15 & 68.61  & 71.60  & 65.73  & 40.12  & 66.80  &\bf72.82  & 70.49  \\ 
  16 & 99.00  & 79.97  & 88.53  & 82.69  & 96.38  & 97.37  &\bf99.32 \\ 
  \hline 
  OA (\%) &89.90  & 85.24  & 81.65  & 76.41  & 87.99  & 84.80  &\bf 92.04  \\
  AA (\%) &94.18  & 91.40  & 88.93  & 83.87  & 93.10  & 89.84  &\bf96.26  \\
  $\kappa$ &88.77  & 83.70  & 79.78  & 73.74  & 86.65  & 83.17  &\bf 91.13 \\
  \bottomrule[1.5pt]
  \end{tabular}}
  \label{Table:Sa comparative}
  \end{table}

% ---------------------------------  GF5 comparative  ---------------------------------
\begin{table}[!t]
  \centering
  \caption{CLASSIFICATION ACCURACY (\%) AND KAPPA MEASURE FOR THE YRE DATASET}
  \setlength{\tabcolsep}{1.2mm}{
  \renewcommand\arraystretch{1.5}
  \begin{tabular}{c|ccccccc}
  \toprule[1.5pt]
  Class&Vit &HybridSN &FC3D &CNNHSI &TwoCNN &SPRN &MCT\\
  \hline  
  1 & 49.78  & 82.35  & 78.84  &\bf92.16  & 70.48  & 86.46  & 85.58  \\ 
  2 & 95.45  &\bf100.00  & 99.93  & 99.12  & 99.96  & 99.90  & 98.45  \\ 
  3 & 55.45  & 61.34  & 72.20  & 75.74  & 85.93  & 79.80  &\bf87.40  \\ 
  4 & 78.22  & 71.04  & 72.78  & 83.22  & 90.44  & 89.67  &\bf93.67  \\ 
  5 & 85.89  & 76.74  & 90.49  & 86.98  & 97.59  &\bf97.81  & 98.26  \\ 
  6 & 79.48  & 81.77  & 83.37  & 85.30  & 82.65  & 82.52  &\bf86.87  \\ 
  7 & 56.38  & 51.53  & 57.30  & 57.89  &\bf63.63  & 57.44  & 60.04  \\ 
  8 & 73.16  & 73.94  & 73.68  & 78.41  & 60.17  & \bf84.03  & 80.34  \\ 
  9 & 75.10  & 85.90  & 86.11  & 86.67  & 82.70  & 86.26  &\bf88.74  \\ 
  10 & 57.06  & 70.82  & 62.15  &\bf90.93  & 72.22  & 88.46  & 90.65  \\ 
  11 & 63.65  & 82.55  & 83.38  & 89.97  & 75.71  & 79.79  &\bf90.80  \\ 
  12 & 72.82  & 76.36  & 79.69  & 77.93  & 73.76  & 77.35  &\bf83.95  \\ 
  13 & 71.27  & 87.47  & 84.49  &\bf92.27  & 87.79  & 90.88  & 91.26  \\ 
  14 & 67.11  & 75.32  & 76.50  & 88.38  & 88.63  &\bf89.95  & 83.75  \\ 
  15 & 59.32  & 64.65  & 74.65  & 85.73  &\bf96.32  & 89.40  & 66.87  \\ 
  16 & 74.27  & 93.02  & 89.89  & 93.73  & 92.20  & 94.84  &\bf96.19  \\ 
  17 & 81.29  & 93.98  & 89.47  & 96.67  & 93.72  &\bf99.26  & 98.40  \\ 
  18 & 44.74  & 58.11  & 62.74  & 65.66  & 67.59  &\bf68.96  & 67.95  \\ 
  19 & 60.53  & 82.89  & 68.25  &\bf88.61  & 68.84  & 80.06  & 82.67  \\ 
  20 & 87.69  &\bf91.28  & 71.79  & 87.69  & 64.00  & 86.77  & 87.69 \\ 
  \hline 
  OA (\%) &75.58  & 76.14  & 80.84  & 84.61  & 87.45  & 88.58  &\bf90.72  \\
  AA (\%) &69.43  & 78.05  & 77.88  & 85.15  & 80.72  & 85.48  &\bf85.98  \\
  $\kappa$ &71.87  & 72.96  & 78.09  & 82.30  & 85.48  & 86.77  &\bf89.26 \\
  \bottomrule[1.5pt]
  \end{tabular}}
  \label{Table:GF5 comparative}
  \end{table}

The results of the comparative experiments are shown in \ref{Table:Sa comparative} and \ref{Table:GF5 comparative}. It can be seen that our method has obtained obvious advantages, and achieved the best or second-best results in classification accuracy in each class, which reflects the superiority and robustness of our method. Under the setting of limited training samples, CNNHSI also achieves excellent classification results due to its lightweight network structure.  The performance of SPRN is also good, which benefits from the design of the spectral partition operation. Limited by the large model size, HybridSN, FC3D, and TwoCNN failed to obtain superior classification results. The Transformer structure is often data-hungry, and it is difficult to directly compete with the CNN structure in small sample scenarios. Therefore, this directly proves the necessity of a well-designed embedding layer for HSI in Transformer network. So, we designed a dual-branch convolution embedding module in a targeted manner, combined with the CM pre-training strategy, to further improve the performance of the model in the case of small samples, and realize the effective use of unlabeled samples. In addition, all the methods can't discriminate the Vinyard untrained class well, which may be limited by the large variability of this land cover, which is also a problem we need to solve in the future. The classification results of each method on the YRE dataset are similar to the Salinas dataset, but the YRE dataset is a large scene dataset, and the classification task is more difficult, so the classification performance of each method is slightly lower than that on the Salinas dataset. It is worth mentioning that TwoCNN achieves good classification results, which may benefit from its spectral feature extraction branch. The classification results of each method on the YRE dataset are similar to the Salinas dataset, but the YRE dataset is a large scene dataset, and the classification task is more complicated. So, the classification performance of each method is slightly lower than that of the Salinas dataset. It is worth mentioning that TwoCNN achieves good classification results, which may benefit from its spectral feature extraction branch.\par

% clsmap for Salinas
\begin{figure}[htbp]
  \centering
  \subfigure[]
  {
      \begin{minipage}[b]{.25\linewidth}
          \centering
          \includegraphics[scale = 0.35]{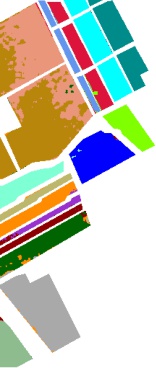}
      \end{minipage}
  }
  \subfigure[]
  {
     \begin{minipage}[b]{.25\linewidth}
          \centering
          \includegraphics[scale = 0.35]{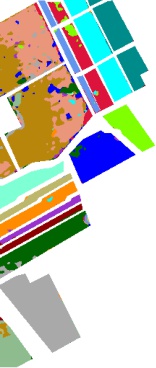}
      \end{minipage}
  }
  \subfigure[]
  {
      \begin{minipage}[b]{.25\linewidth}
          \centering
          \includegraphics[scale = 0.35]{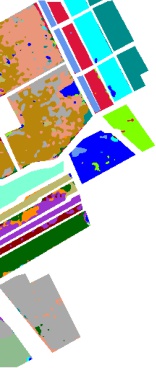}
      \end{minipage}
  }
  
  \subfigure[]
  {
     \begin{minipage}[b]{.25\linewidth}
          \centering
          \includegraphics[scale = 0.35]{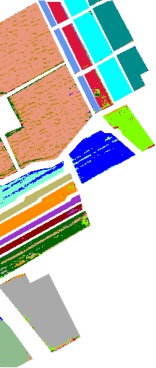}
      \end{minipage}
  }
  \subfigure[]
  {
      \begin{minipage}[b]{.25\linewidth}
          \centering
          \includegraphics[scale = 0.35]{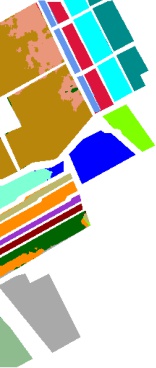}
      \end{minipage}
  }
  \subfigure[]
  {
      \begin{minipage}[b]{.25\linewidth}
          \centering
          \includegraphics[scale = 0.35]{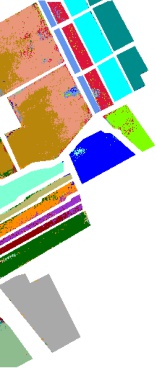}
      \end{minipage}
  }
  
  \subfigure[]
  {
     \begin{minipage}[b]{.25\linewidth}
          \centering
          \includegraphics[scale = 0.35]{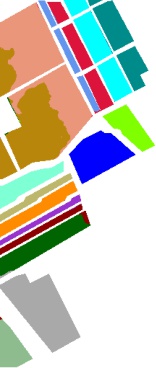}
      \end{minipage}
  }
  \subfigure[]
  {
      \begin{minipage}[b]{.25\linewidth}
          \centering
          \includegraphics[scale = 0.35]{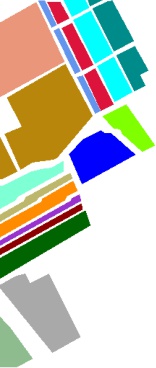}
      \end{minipage}
  }
  \subfigure[]
  {
      \begin{minipage}[b]{.25\linewidth}
          \centering
          \includegraphics[scale = 0.34]{cls_map/Sa.png}
      \end{minipage}
  }
  
  \caption{Salinas dataset. Classification maps obtained by (a) CNNHSI(89.90\%), (b) 3D-2D-CNN(85.24\%), (c) 3D-CNN(81.65\%), (d) Two-CNN(76.41\%), (e) SPRN(87.99\%), (f) Vit(84.80\%), (g) Ours(92.04\%),and (h) Ground-truth and (j) False-color image}
  \label{fig:Saclsmap}
  \end{figure}

% clsmap for GF5
\begin{figure}[htbp]

  \centering
  
  \subfigure[]
  {
      \begin{minipage}[b]{.25\linewidth}
          \centering
\includegraphics[scale = 0.07]{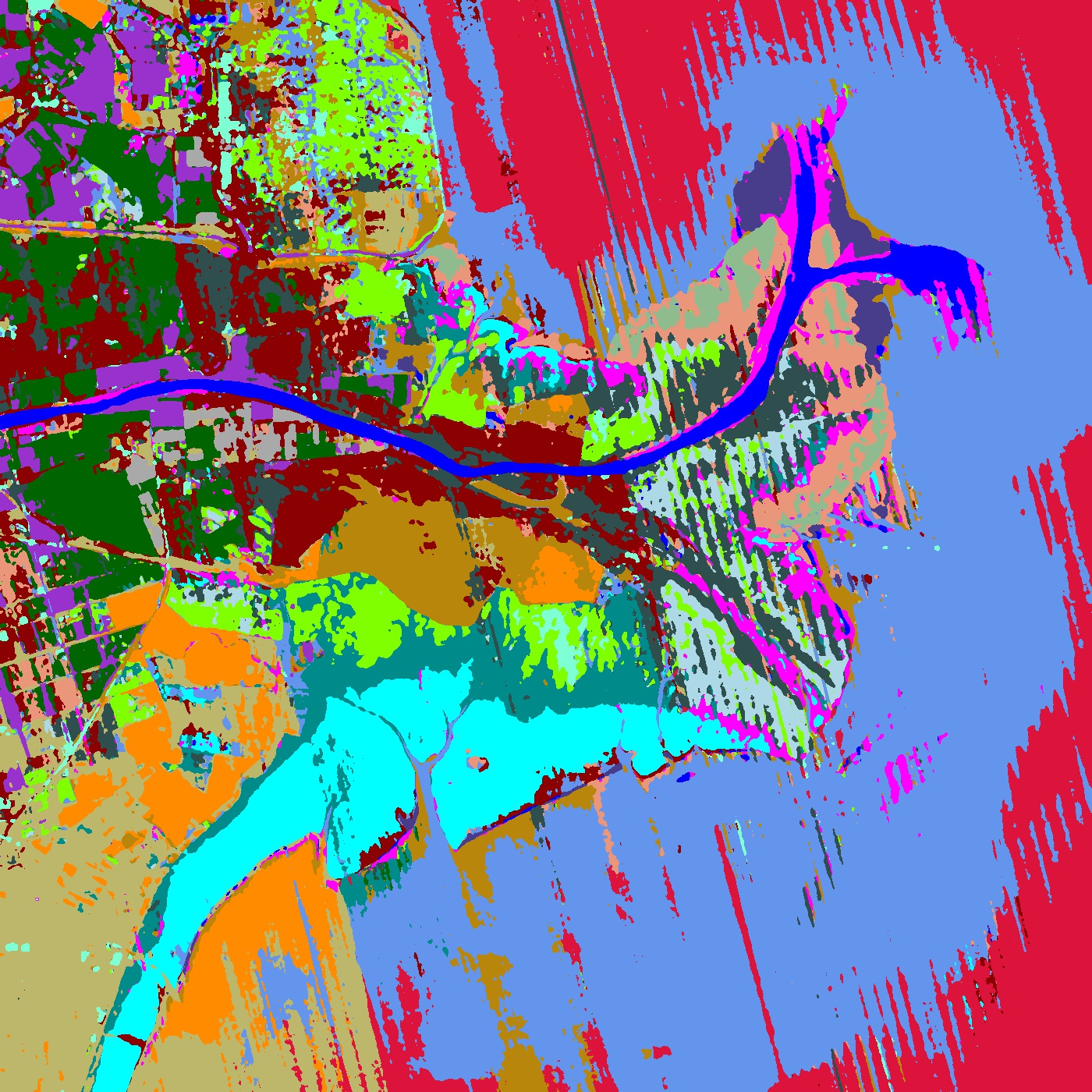}
      \end{minipage}
  }
  \subfigure[]
  {
     \begin{minipage}[b]{.25\linewidth}
          \centering
          \includegraphics[scale = 0.07]{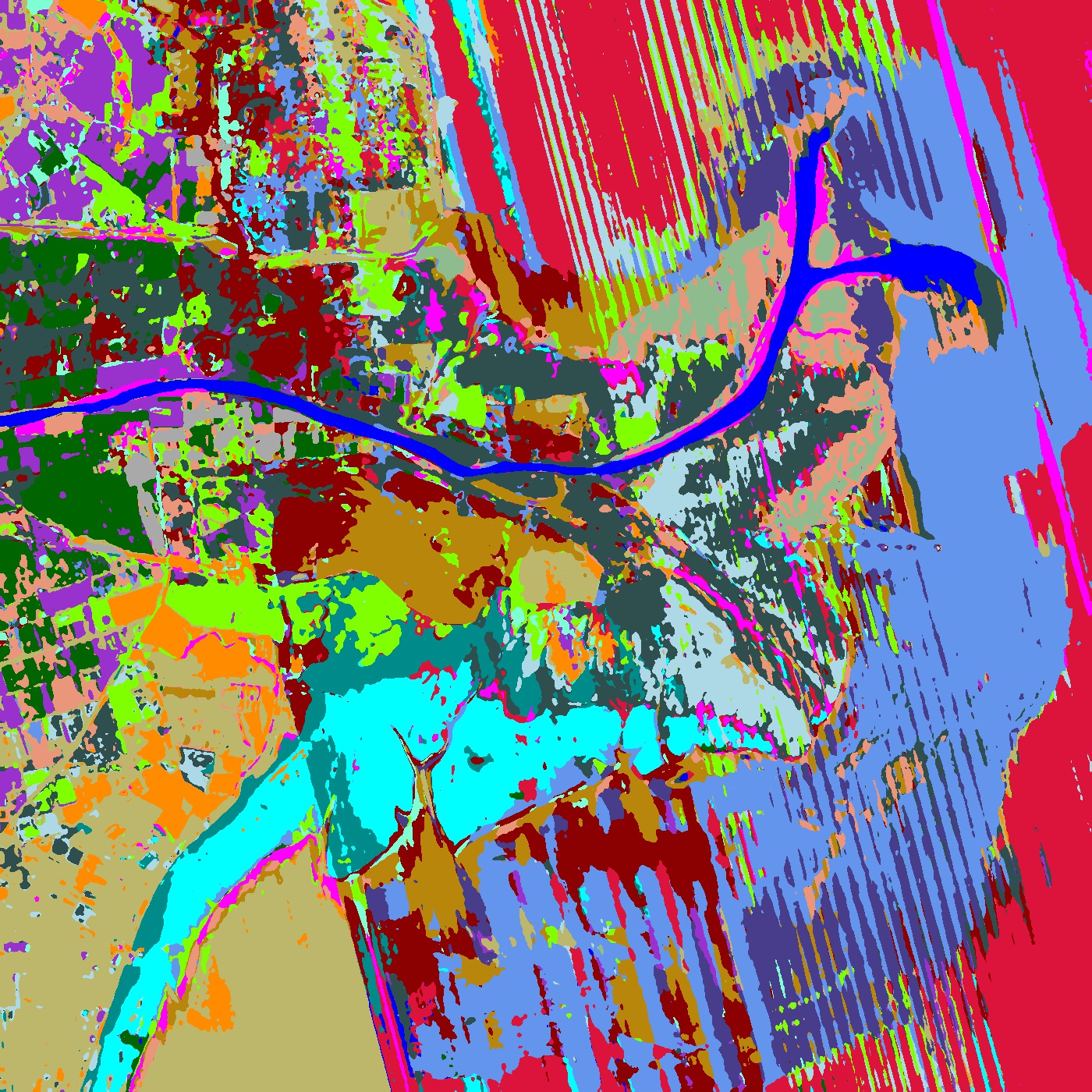}
      \end{minipage}
  }
  \subfigure[]
  {
      \begin{minipage}[b]{.25\linewidth}
          \centering
          \includegraphics[scale = 0.07]{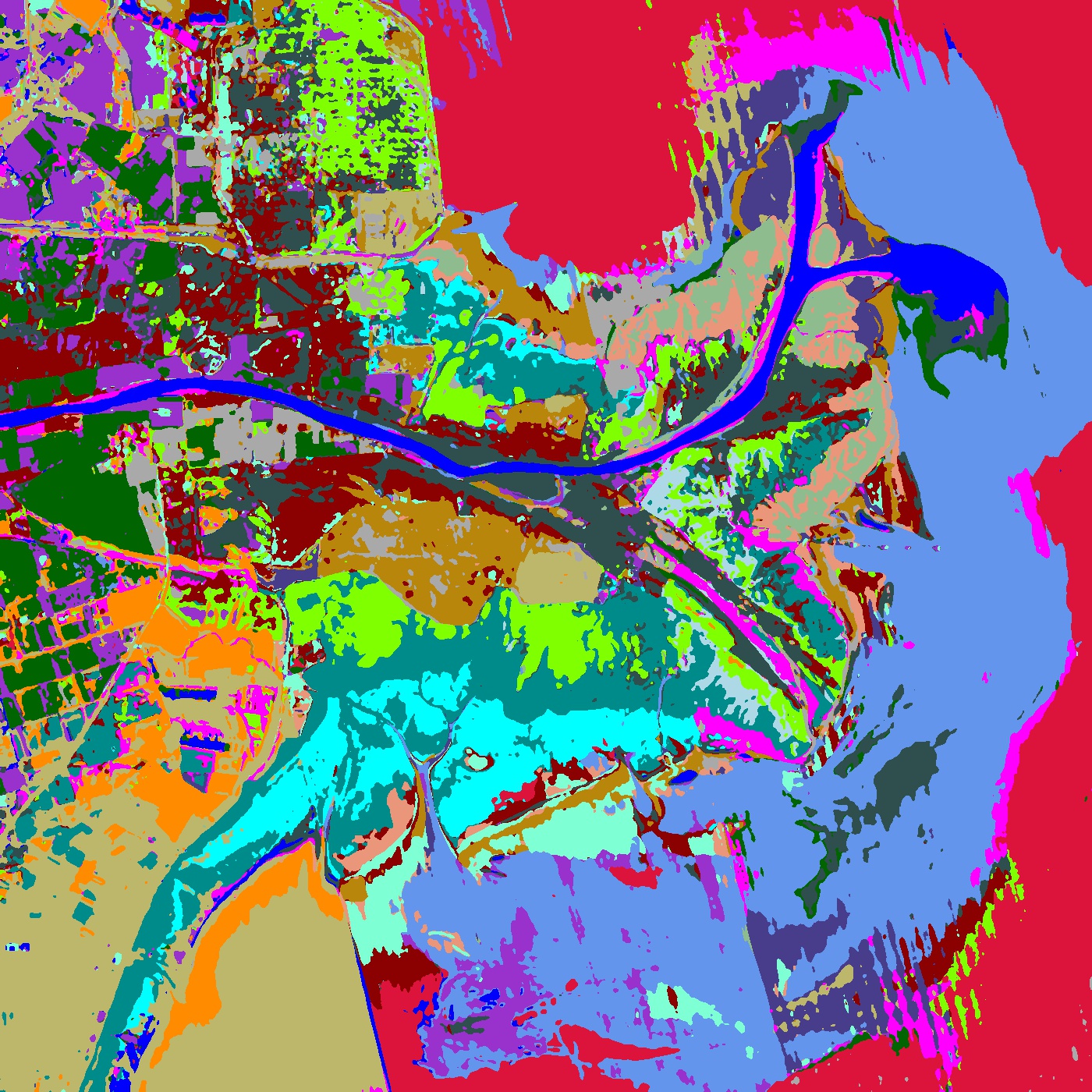}
      \end{minipage}
  }
  \subfigure[]
  {
     \begin{minipage}[b]{.25\linewidth}
          \centering
          \includegraphics[scale = 0.07]{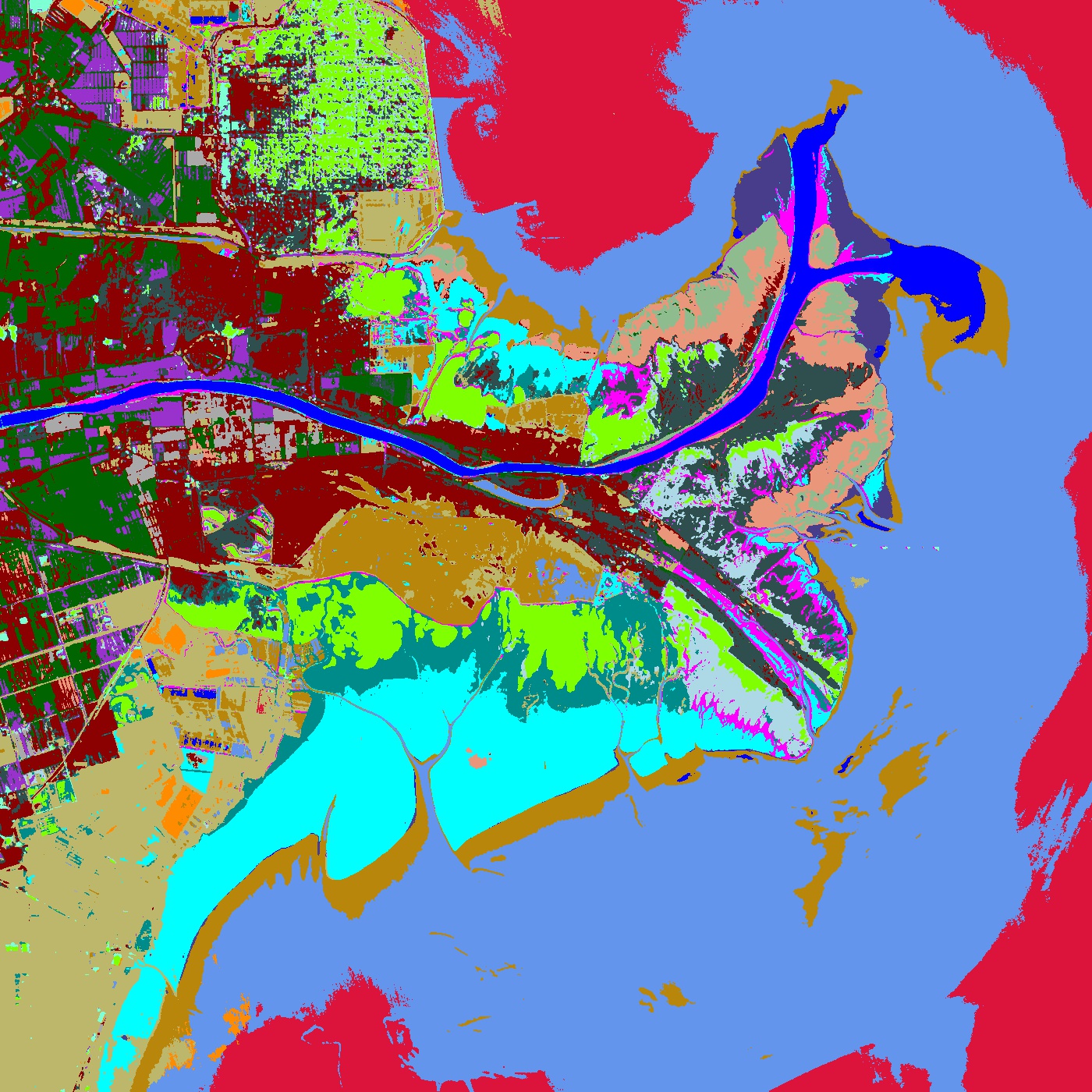}
      \end{minipage}
  }
  \subfigure[]
  {
      \begin{minipage}[b]{.25\linewidth}
          \centering
          \includegraphics[scale = 0.07]{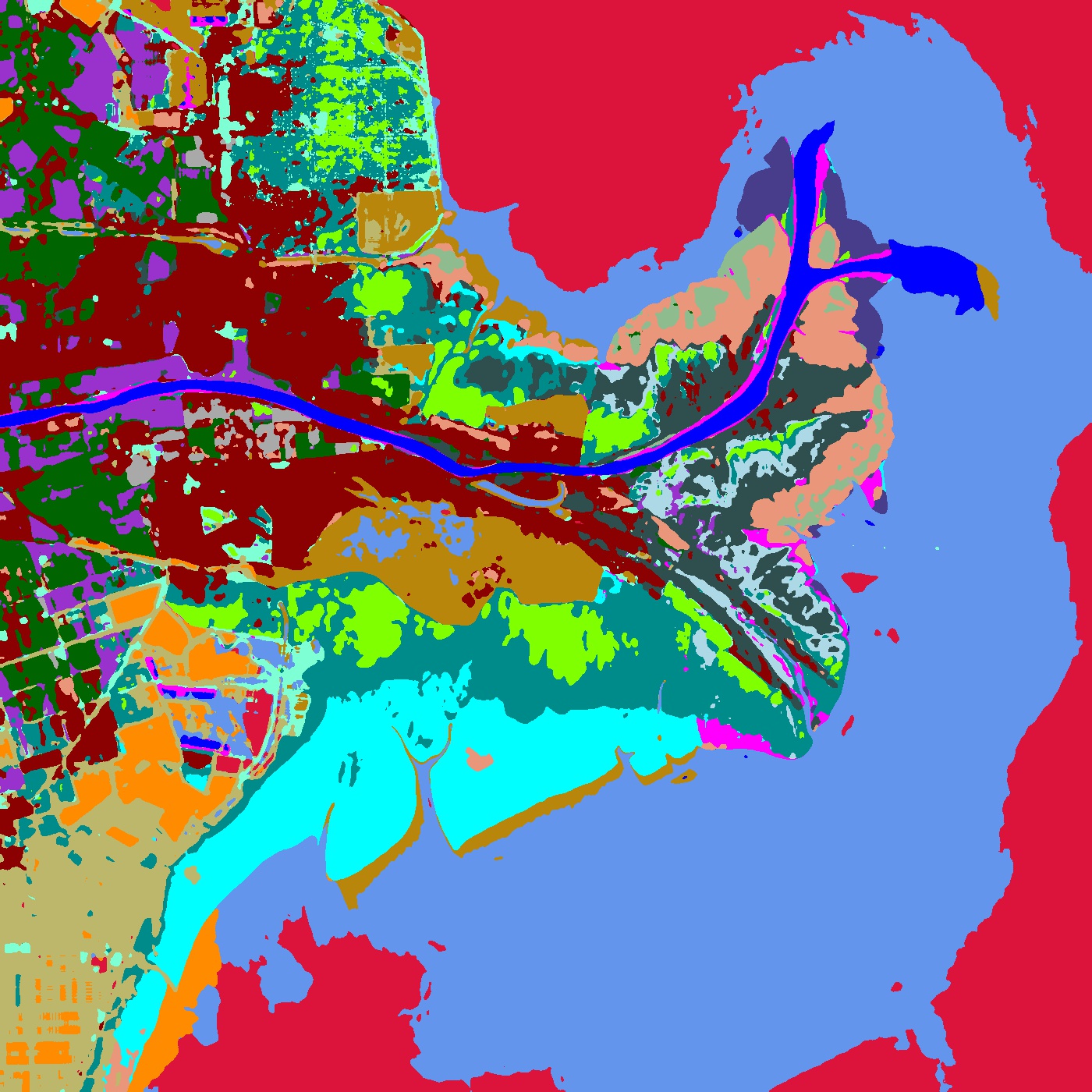}
      \end{minipage}
  }
  \subfigure[]
  {
      \begin{minipage}[b]{.25\linewidth}
          \centering
          \includegraphics[scale = 0.07]{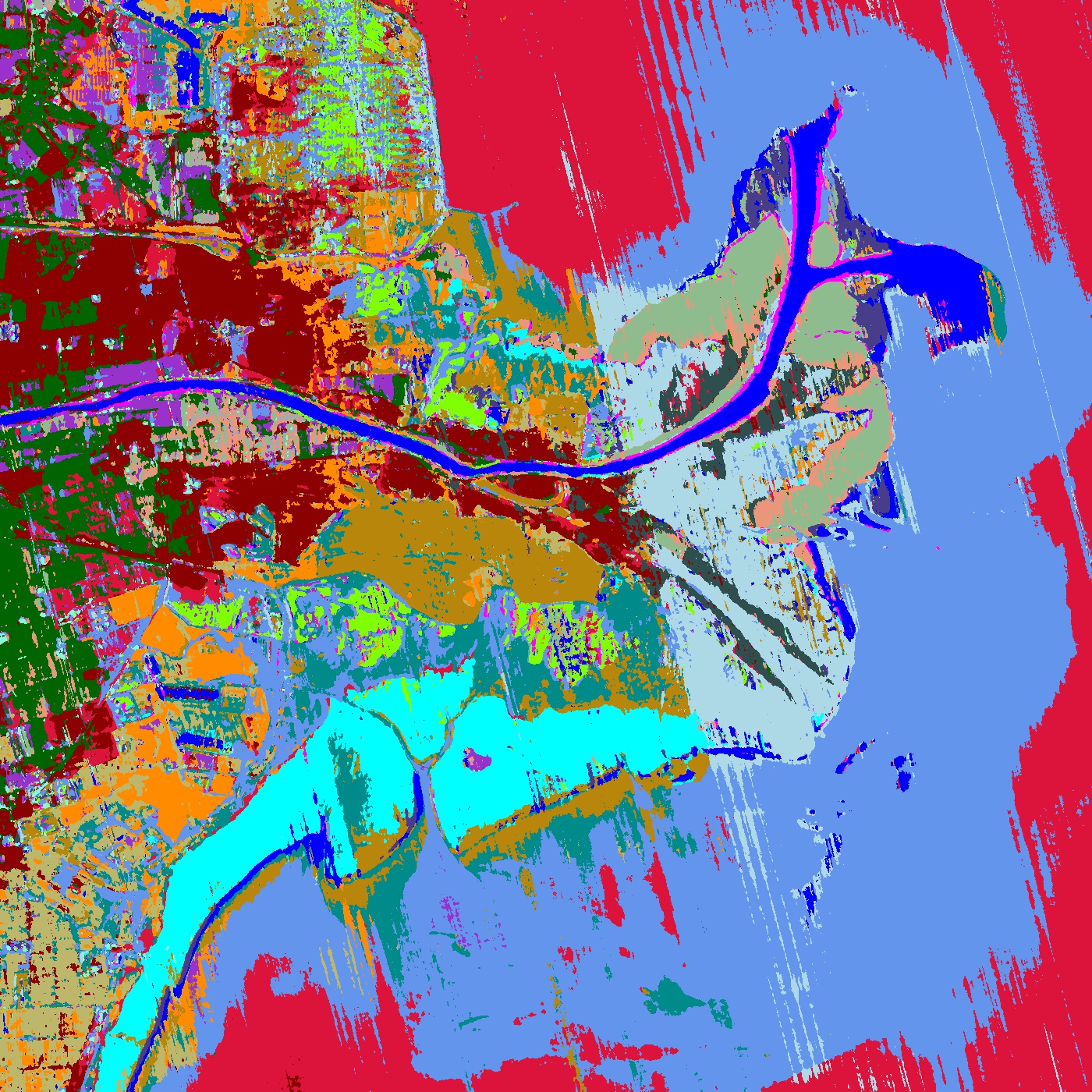}
      \end{minipage}
  }
  \subfigure[]
  {
     \begin{minipage}[b]{.25\linewidth}
          \centering
          \includegraphics[scale = 0.07]{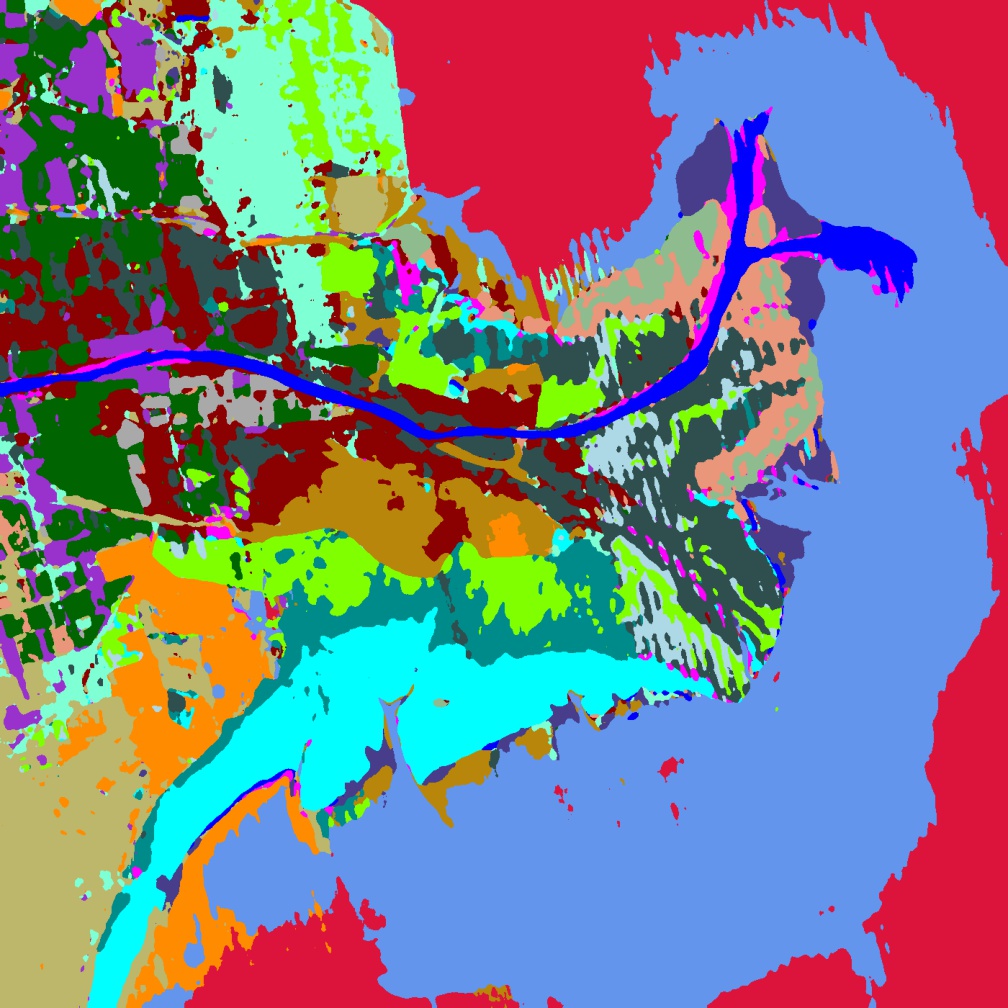}
      \end{minipage}
  }
  \subfigure[]
  {
      \begin{minipage}[b]{.25\linewidth}
          \centering
          \includegraphics[scale = 0.07]{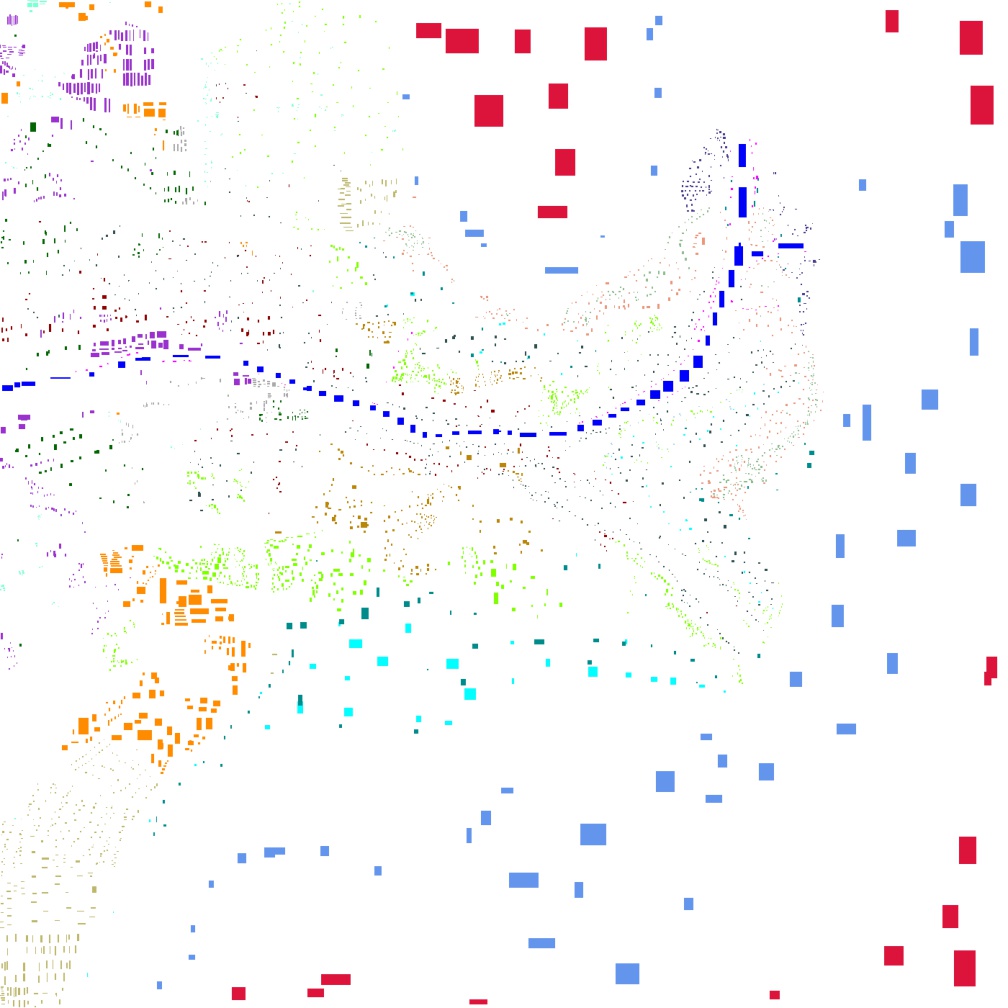} 
      \end{minipage}
  }
  \subfigure[]
  {
      \begin{minipage}[b]{.25\linewidth}
          \centering
          \includegraphics[scale = 0.23]{cls_map/False_GF5.png} 
      \end{minipage}
  }
  
  \caption{YRE dataset. Classification maps obtained by (a) CNNHSI(89.90\%), (b) 3D-2D-CNN(85.24\%), (c) 3D-CNN(81.65\%), (d) Two-CNN(76.41\%), (e) SPRN(87.99\%), (f) Vit(84.80\%), (g) Ours(92.04\%), (h) Ground-truth, and (j) False-color image}
  \label{fig:GF5clsmap}
  \end{figure}

Besides, considering that the YRE dataset did not completely label all samples, to fully display the classification performance between comparison methods, We use partial prediction and full prediction in the generation of classification results for the Salinas and YRE datasets, respectively.That is, for the Salinas dataset, we only visualize labeled samples, and for YRE, we make predictions for the entire dataset. The classification map of Salinas is shown in [Figure]. It can be seen that the classification results of our method are almost consistent with the Groundtruth in most classes, while other methods have different numbers of classification errors in almost every land covers. This intuitively reflects the stability of our method. The classification map of YRE is shown in [Figure]. Similarly, our proposed method restores the distribution of surface objects well, but there is a smearing phenomenon. In contrast, although the classification accuracy of SPRN is slightly lower, its classification map appears smoother. This is also the direction of improvement that we need in the future.

\subsection{Ablation Study}
In this section we conduct comprehensive ablation experiments on the multi-scale convolutional embedding module and the center mask-reconstruction pre-training pretask (PTP). The experimental results are shown in table \ref{Table:ablation}. The experimental results compared with the Vit model can intuitively demonstrate the effectiveness of our proposed MCE module. On this basis, we further demonstrate the effectiveness of the IIE branch in the MCE module. For the convenience of description, we name the MCT model after removing the IIE branch as 3DCT. The experimental results on the Salinas and YRE datasets show that the accuracy of MCT is 0.47\% and 2.12\% higher than that of 3DCT, respectively. \par

Finally, various ablation experiments are performed on our proposed pre-training pretask. The experimental results on the Salinas and YRE datasets show that the model fine-tuned on pre-trained model (PTM) outperforms the model without pre-training by 2.97\% and 2.2\%, respectively. In addition, we do additional experiments on the 3DSPCT model, fine-tuning 3DSPCT on PTM of MCT (removing the parameters of the IIE module). The experimental results prove that no matter fine-tuning on PTM of 3DSPCT or MCT, competitive experimental results can be obtained. This undoubtedly proves the superiority and robustness of our pre-training pretask.

% module ablation
\begin{table}[!t]
\centering
\caption{ABLATION STUDY RESULTS TOWARD THE MCE MODULE AND CMR PRETRAINING PRETASK ON SALINAS AND YRE DATASETS}
\setlength{\tabcolsep}{2mm}{
\renewcommand\arraystretch{1.5}
\begin{tabular}{c|cc|cc|ccc}
    \toprule[1.5pt]
    \multirow{2}{*}{Dataset} & \multicolumn{2}{c|}{Module} & \multicolumn{2}{c|}{Pretrain Model} & \multicolumn{3}{c}{Metric} \\
    \cline{2-8} & 3DSPCE &IIE &full &partial & OA (\%) & AA (\%) & $\kappa$ \\
    \hline
    
    Salinas &\checkmark &\checkmark &\XSolidBrush &\XSolidBrush   &\bf89.12 &\bf95.03 &\bf87.92 \\
    Salinas &\checkmark &\XSolidBrush &\XSolidBrush &\XSolidBrush   &88.60 &85.42 &86.46 \\
    \cline{2-8}
   Salinas &\checkmark &\checkmark &\checkmark &\XSolidBrush   &\bf92.04 &\bf96.26 &\bf91.13 \\     
   Salinas &\checkmark &\XSolidBrush &\checkmark &\XSolidBrush   &89.99 &95.40 &88.87 \\   
   Salinas &\checkmark &\XSolidBrush &\XSolidBrush &\checkmark   &89.66 &95.85 &88.54 \\

    \hline
    
    YRE &\checkmark &\checkmark &\XSolidBrush &\XSolidBrush   &\bf88.64 &\bf85.34 &\bf86.88 \\
    YRE &\checkmark &\XSolidBrush &\XSolidBrush &\XSolidBrush   &86.24 &82.19 &84.15 \\
    \cline{2-8}
   YRE &\checkmark &\checkmark &\checkmark &\XSolidBrush   &\bf90.72 &\bf85.98 &\bf89.26 \\     
   YRE &\checkmark &\XSolidBrush &\checkmark &\XSolidBrush   &89.01 &84.11 &87.30 \\   
   YRE &\checkmark &\XSolidBrush &\XSolidBrush &\checkmark   &90.51 &\bf86.01 &89.01 \\

\bottomrule[1.5pt]
\end{tabular}}
\label{Table:ablation}
\end{table}

\section{Conclusion}
In this article, we propose a reconstruction-based pre-training pretask and lightweight backbone network for HSI. We creatively propose a multi-scale convolutional embedding module to extract spatial spectral information at different scales with 3D convolution and linear embedding using spectral segmentation strategy, respectively, to obtain more discriminative embeddings. The Transformer module is used to model the global relationship between surface object representations. Additionally, for few-shot scenarios, we propose a new unsupervised pre-training pretask suitable for hyperspectral classification tasks. Through the mask and reconstruction process of the token generated based on the central pixel vector, the model can better learn the relationship between the neighboring surface objects and the central surface object during the pre-training process, and can provide more stable experimental results and model performance. We conducted a series of comparative experiments and ablation experiments on two public datasets, Salinas and YRE, to demonstrate the effectiveness and superiority of our proposed method.


% \section*{Acknowledgment}

% The authors would like to thank myself.

\bibliographystyle{ieeetr}
\bibliography{reference}

\end{document}